\documentclass[conference,a4paper]{IEEEtran}
\IEEEoverridecommandlockouts
\usepackage{cite}
\let\citep\cite
\usepackage{amsmath,amssymb,amsfonts}
\usepackage{graphicx}
\usepackage{textcomp}
\usepackage{xcolor}
\usepackage{booktabs,tabularx,array,longtable}
\usepackage{enumitem}
\usepackage{setspace}
\usepackage{xurl}
\usepackage{hyperref}
\hypersetup{hidelinks,pdftitle={SAFE-AI: Evidence-linked Privacy Assessment before University AI Deployment}}

\setlist{itemsep=3pt,topsep=4pt,parsep=1pt,leftmargin=*}
\newcolumntype{P}[1]{>{\raggedright\arraybackslash}p{#1}}
\newcolumntype{Y}{>{\raggedright\arraybackslash}X}
\newcommand{\crit}[1]{\textsf{C#1}}
\newcommand{\rqone}{What privacy risks arise from the use of AI systems in university teaching, research and administration?}
\newcommand{\rqtwo}{What do existing privacy impact assessment approaches cover for AI-specific privacy risks, and what work remains for a university deployment?}
\newcommand{\rqthree}{What additional criteria should be incorporated into an AI-specific privacy risk assessment framework?}
\newcommand{\rqfour}{How can the proposed framework support universities in determining whether an AI application presents acceptable privacy risk?}

\begin{document}

\title{SAFE-AI: Evidence-linked Privacy Assessment\\before University AI Deployment}

\author{\IEEEauthorblockN{1\textsuperscript{st} Abhishek Sebastian}
\IEEEauthorblockA{\textit{The University of Sydney}\\
Sydney, Australia\\
aseb0376@uni.sydney.edu.au}
\and
\IEEEauthorblockN{2\textsuperscript{nd} Sree Lakshmi Girisan}
\IEEEauthorblockA{\textit{The University of Sydney}\\
Sydney, Australia\\
sgir0712@uni.sydney.edu.au}
\and
\IEEEauthorblockN{3\textsuperscript{rd} Deepali Vinayak}
\IEEEauthorblockA{\textit{The University of Sydney}\\
Sydney, Australia\\
dvin0735@uni.sydney.edu.au}
\and
\IEEEauthorblockN{4\textsuperscript{th} Jai Padmakar Telang}
\IEEEauthorblockA{\textit{The University of Sydney}\\
Sydney, Australia\\
jtel0742@uni.sydney.edu.au}
\and
\IEEEauthorblockN{5\textsuperscript{th} Devadatta Bhat}
\IEEEauthorblockA{\textit{The University of Sydney}\\
Sydney, Australia\\
dbha5912@uni.sydney.edu.au}
}

\maketitle

\begin{abstract}
Universities are adopting artificial intelligence (AI) tools for teaching, research and administration faster than they can check what those tools do with personal information. Existing privacy impact assessments ask many of the right questions but rarely say what to decide when an answer is missing, or how to treat a serious but unlikely harm. This paper develops SAFE-AI, the Structured Assessment Framework for Educational AI, from a targeted review of research, Australian regulatory guidance and public university policies. Three documented problems shape it: students often cannot refuse a required tool, removing names does not remove sensitive meaning, and a reassuring claim about one part of a service can hide risk in another. SAFE-AI combines a twenty-question checklist, five mandatory pass/fail gates, simple harm ratings and a signed decision record; missing evidence pauses a decision, and severe harm can never reach standard review. Applied to ANU's Project Lyrebird and the University of Sydney's ChatGPT Edu using public pages only, it separated data storage from processing location, held Lyrebird pending study ethics evidence and gave ChatGPT Edu a conditional approval with named actions. On eight realistic multi-document dossiers, five persona-based AI assessors using SAFE-AI caught every planted blocking problem (30/30, against 28/30 for a full OAIC-based assessment), matched the OAIC method on a low-risk conditional case, though it never fully approved the clean dossier, and agreed more often. Testing with human assessors is the necessary next step.
\end{abstract}

\begin{IEEEkeywords}
privacy impact assessment, university AI, AI governance, learning analytics, Australian higher education, residual risk
\end{IEEEkeywords}

\section{Introduction}
\label{sec:intro}
Consider a second-year university student. In one semester their unit introduces an AI tutor that reads their draft essays, the counselling service trials an AI note-taker for their appointments, and a researcher invites them to an interview that an AI tool will transcribe and summarise. Each decision was made by a different part of the university, and each looked small. Together they change who can see the student's information, what can be guessed about them, where their words are processed and how long copies survive. The question this paper asks is therefore not whether AI is useful in general, but whether a specific AI deployment is justified, supported by evidence and properly governed before personal information enters it.

The question is urgent. Sydney and Monash publish AI principles and policies \citep{sydneyprinciples,monashpolicy}. Adoption is also widening: Sydney plans to give all students access to ChatGPT Edu by Semester~1~2027, with optional \emph{agents} (AI that can take actions on a user's behalf) and \emph{connectors} (links that let the AI read other systems, such as email or student records) to be reviewed separately \citep{sydneyedu}. Each new feature changes who can retrieve records and where prompts are processed. Universities do not lack governance. What they lack is a repeatable way to turn that governance into a decision about one particular workflow. A tool approved for public course material is not thereby approved for health disclosures, unpublished interview transcripts or decisions about individual students.

Privacy is also broader than keeping data secret. Unwanted observation, unjustified guesses about a person, reuse of data for a new purpose and the inability to challenge a wrong record can all cause harm even when encryption works perfectly. Students may have little real choice about a required platform; research participants may have agreed to an interview but not to an outside AI company processing it; staff may be monitored beyond what they expect. Learning-analytics research has long recognised these power imbalances \citep{slade2013,pardo2014}, and security research adds ways in which AI models themselves can leak information \citep{shokri2017,carlini2021}.

This paper proposes SAFE-AI, the Structured Assessment Framework for Educational AI. Its aim is to help a university decide, before deployment and whenever something material changes, whether an AI application presents acceptable privacy risk. We ask four research questions:
\begin{enumerate}[label=\textbf{RQ\arabic*.},leftmargin=17mm,itemsep=4pt,topsep=6pt]
\item \rqone
\item \rqtwo
\item \rqthree
\item \rqfour
\end{enumerate}

The paper makes three contributions:
\begin{itemize}
\item a twenty-criterion checklist, derived traceably from research and guidance, that asks for evidence about a specific configuration rather than a product in general;
\item a decision rule with five mandatory gates, in which missing evidence pauses a decision and a severe harm can never reach standard review;
\item a documentary application to two real public university deployments, and a comparative test against a full OAIC-based assessment on realistic case dossiers using persona-based simulated assessors.
\end{itemize}

\section{Background and Problem Context}
\label{sec:context}
\subsection{Key concepts}
A \emph{privacy risk} is a plausible chain of events through which handling data harms a person or group. SAFE-AI records four parts of each chain: the information involved, the event that exposes or misuses it, the people affected and the resulting harm. An \emph{inherent} rating describes the risk before any safeguards are applied; a \emph{residual} rating describes what remains after safeguards that are backed by evidence. A promised safeguard with no evidence does not lower the rating. A \emph{privacy impact assessment} (PIA) is a structured review of a project's privacy effects and responses. It covers one project at one time, and it does not certify a product forever \citep{oaicpia}.

SAFE-AI assesses a \emph{deployment configuration}: one purpose, one population, its data sources, the exact model or service version, the \emph{tenant}, connected systems and permitted uses. It matters because AI services handle data in several different ways. \emph{Training} changes a model using data; \emph{inference} is the model processing a request to produce an answer; \emph{retrieval-augmented generation} (RAG) feeds retrieved documents, such as student files, to the model. A service can promise never to train on prompts yet still process sensitive prompts overseas, keep logs, or retrieve another person's record through RAG. And removing names does not stop people being recognised from context.

Which laws apply must be checked, not assumed: most universities fall outside the federal Privacy Act's general coverage, with exceptions including ANU, and state rules such as NSW privacy law, with the HRIP Act for health information, can apply instead \citep{oaicscope,sydneynotice,hripact}. From December 2026, Privacy Act amendments also require covered entities, including ANU, to state in privacy policies the kinds of significant decisions made or substantially assisted by computer programs \citep{polaa2024}. SAFE-AI therefore keeps a jurisdiction-specific compliance record alongside ethics approvals, policies and contracts.

\subsection{Three documented privacy anomalies}
We use \emph{anomaly} to mean a gap between an apparent protection, or a legitimate aim, and the privacy consequences that remain. Each example comes from published research or a university's own public statements. None suggests that a named university has broken the law.

\subsubsection{A1: required participation can undermine meaningful choice}
Online exam proctoring shows the tension between assessment integrity and students' control over being watched. Balash et al.~\cite{balash2021} surveyed 102 participants and documented privacy and security concerns under institutional requirements, and later stakeholder research examines how responsibility for proctoring privacy is perceived in higher education \citep{mutimukwe2026}. Products differ, but a notice or consent button is a weak safeguard when refusing means failing the assessment. So the real question is whether the watching is necessary, and whether a student can sit the exam another way. Where the answer is no, students expose their homes and can be treated with suspicion because an algorithm flagged them. Recording that terms were accepted is not enough; the assessment has to test necessity, the alternative and the route to challenge a flag (criteria \crit{01}, \crit{03} and \crit{14}).

\subsubsection{A2: removing identifiers can leave sensitive meaning exposed}
Mireshghallah et al.~\cite{mireshghallah2024} found personal disclosures throughout real chatbot conversations and showed that tools which detect names and numbers miss sensitive topics. Their users were not university students, but the mechanism carries over. Asking an AI to edit, translate or summarise text can reveal intimate context. Research interviews and student-support notes carry the same exposure after names are removed. In a small study, one distinctive story can identify a participant. Participants who learn that an outside company's model reads their words may withdraw or hold back. Summaries, logs and \emph{embeddings} (Appendix~\ref{app:glossary} defines these terms) can also survive after the transcript itself is deleted. ANU's Project Lyrebird policy is a public example of handling this through study-specific consent and processing arrangements \citep{anulyrebird}.

\subsubsection{A3: one reassurance can hide a different boundary}
The University of Sydney's ChatGPT Edu pages state that data is stored in Australia at rest but that prompts may be processed transiently outside Australia, and note administrator access and separately assessed optional features \cite{sydneyedu,sydneyedunews}. Knowing where data is stored tells you nothing about where it is processed. A promise not to train on conversations says nothing about who can read the ones that are kept. If a connector is enabled or administrator log access widened, counselling or misconduct notes could reach an overseas processing path while the storage promise remains true.

\section{Literature Review and Related Work}
\label{sec:lit}
This section traces four bodies of work: educational privacy research, which predates generative AI; impact-assessment methods in Australia and abroad; technical research on AI leakage; and current Australian university policy.

\subsection{Educational privacy predates generative AI}
Learning-analytics research established the ethical baseline for student data. Slade and Prinsloo~\cite{slade2013} placed analytics inside institutional power relationships and learner agency; Pardo and Siemens~\cite{pardo2014} set out ethical and privacy principles for educational data. DELICATE turned these concerns into a practical checklist, and LEA in Private built a privacy framework for a learning-analytics toolbox \citep{drachsler2016,steiner2016}. Cormack~\cite{cormack2016} distinguished discovering patterns from intervening with individuals, a distinction that matters when an AI prediction enters a student-support workflow. Hakimi et al.~\cite{hakimi2021} reviewed the ethics of digital trace data in education and catalogued proposed responses, including holistic impact assessments.

These frameworks took privacy seriously. They were written before prompts, embeddings and fast-changing AI features, though, so they say nothing about what evidence to ask for. An AI checklist that dropped their attention to power and intervention would lose essential context. The concern is current: Liu and Khalil~\cite{liu2023} found privacy issues across 47 learning-analytics studies, Wang et al.~\cite{wang2025} report student concerns about generative AI, and Cormack and Reeve~\cite{cormack2022} warn that wellbeing interventions can create sensitive labels.

\subsection{Impact assessment: strong foundations, weak validation}
Reviews by Wairimu et al.~\cite{wairimu2024} and Iwaya et al.~\cite{iwaya2024} show that having a PIA method is not the same as showing it works in practice, and Stahl et al.~\cite{stahl2023} map a large but uneven landscape of AI impact assessments. In Australia, the OAIC's PIA guide covers project description, consultation, data-flow mapping, impact analysis and risk management \citep{oaicpia}. Its AI guidance and checklist already ask about inputs, outputs, supplier access and environments \citep{oaicuse,oaicchecklist}, and its training guidance notes that public availability does not authorise every reuse of personal information \citep{oaictrain}. We therefore compare SAFE-AI with this AI-specific guidance, since a generic PIA would be too easy a baseline.

International practice points the same way. Under Europe's General Data Protection Regulation (GDPR), a data protection impact assessment (DPIA) is mandatory for high-risk processing \cite{gdpr}, and the EU AI Act requires public-body deployers of high-risk AI, which includes admissions and exam monitoring, to add a fundamental-rights impact assessment \cite{euaiact}. Rintam{\"a}ki et al.~\cite{rintamaki2026} map where the two overlap and argue that the AI assessment should reuse DPIA work rather than duplicate it; an early tool design pursues exactly that reuse \citep{rintamaki2024}. Kasirzadeh and Clifford~\cite{kasirzadeh2021} find that DPIA guidance rarely engages with fairness in practice, a reminder that inference harms can slip through a data-protection review. Canada's Algorithmic Impact Assessment takes a numerical route: 106 questions produce an impact level from I to IV, and strong mitigation can lower the score \citep{canadaaia}. SAFE-AI shares the aim of extending existing assessments, but differs on that last point: mitigation cannot offset a failed mandatory gate or a severe harm.

The US NIST AI Risk Management Framework and its generative-AI profile organise AI risk across the lifecycle \citep{nist2023,nist2024}, and the Cloud Security Alliance publishes control catalogues with vendor questionnaires \citep{csaccm2026,csaaicm2026}. Neither tells a university when to say yes.

\subsection{Technical evidence needs an explicit threat model}
Security research shows that AI models can leak training data. A \emph{membership inference attack} tries to tell whether a person's record was used in training; an \emph{extraction attack} tries to make the model reproduce training text. Both are possible \citep{shokri2017,carlini2021}, but results depend heavily on the model, its training and the attacker's access: rigorous evaluation needs low false-positive rates \citep{carlini2022}, many attacks on pre-trained language models perform near chance \citep{duan2024}, and stronger attacks remain unreliable for individual decisions \citep{hayes2025}; Wu and Cao~\cite{wu2025}, a preprint, is used only as a survey guide. A weak attack that fails proves little, and a successful laboratory attack does not tell us how often a campus system leaks. SAFE-AI therefore treats model leakage as its own evidence question, separate from the far more ordinary risk of a prompt or retrieved file reaching the wrong person.

Vendor policies add a further layer: King et al.~\cite{king2025} found default reuse of chat data and disclosure gaps across six frontier developers, so procurement must check the actual product tier, contract and \emph{subprocessors} (companies the vendor passes data to).

\subsection{Australian university policies and the remaining gap}
Australian universities already govern AI. The University of Sydney ties AI use in assessment to unit rules and reviews ChatGPT Edu feature by feature \citep{sydneyuse,sydneyedu}; Monash's AI Operations Policy requires privacy and security assessment and vendor scrutiny \citep{monashpolicy}; ANU requires approved tools, consents and data-custodian involvement, and documents Lyrebird's processing and withdrawal arrangements \citep{anuguidance,anulyrebird}. Chan~\cite{chan2023} proposes a broader AI policy framework for teaching and learning.

The literature agrees that assessment should target one deployment and reuse existing data-protection work. It disagrees on whether mitigation may offset risk, as Canada's scored tool allows, and on how far membership-inference attacks threaten deployed systems. Across all four bodies of work the remaining gap is practical integration: a record that links a specific harm to evidence, controls, decision authority and reassessment, with explicit rules for missing evidence and severe harm. Whether that is more useful than a competent existing PIA is an empirical question. We test part of it in Section~\ref{sec:panel}.

\section{Research Method}
\label{sec:method}
\subsection{Study design and source selection}
This study follows design-science research, in which a purpose-built artefact is developed and then demonstrated and tested on realistic cases \cite{hevner2004}. The literature review is a targeted narrative review: because the aim was to build an assessment tool from the most relevant evidence, sources were selected for what they contribute to the design. Google Scholar, regulator, publisher and university sites were searched up to 10~October~2026. Searches combined AI with privacy impact assessment, DPIA, algorithmic impact assessment, proctoring, learning analytics, admissions, membership inference, prompt logging, NIST AI RMF and OAIC guidance; 56 sources were included (29 scholarly, 17 regulatory, legislative or standards, 8 university, 2 vendor policy). A source was included if it addressed at least one research question; vendor marketing was never treated as evidence, and AI policies from the University of Sydney, Monash and ANU were chosen as contrasting governance examples, not as a sample of all Australian universities; the two worked cases come from Sydney and ANU.

\subsection{Analysis, construction and evaluation}
Each source was read for the privacy mechanism it described, the protection it proposed and the limits on applying it to universities. Findings were mapped to three questions (what can go wrong, what evidence an assessor needs, and what action follows) and grouped under five themes that became the gates G1--G5. Merging near-duplicates and splitting questions that needed different evidence gave twenty criteria, whose sources are traced in Appendix~\ref{app:checklist}. Existing approaches were compared by what each contributes and what work remains, not by keyword counts.

SAFE-AI was then applied to two real deployments, ANU Project Lyrebird and University of Sydney ChatGPT Edu, using public pages only (Section~\ref{sec:eval}), and compared with a full OAIC-based assessment on realistic case dossiers using a simulated assessor panel (Section~\ref{sec:panel}). No interviews, private records or live systems were used, so no human-research ethics approval was required. The two cases rely only on published pages, the test dossiers are fictional and contain no real personal data, and simulated assessors were used instead of staff so that no one's time or professional judgement was recorded.

\section{Findings: Risks and Existing Approaches}
\label{sec:findings}
This section answers RQ1 and RQ2.

\subsection[RQ1: University AI privacy risks]{RQ1: \rqone}
The review identifies privacy risks at three connected levels. First, \emph{purpose}: collection and use can go beyond what teaching or research requires, through intrusive monitoring, broad collection of personal stories or reuse of records for new aims. Second, \emph{processing}: the AI workflow itself can widen disclosure through prompts, training, retrieval, outputs, logs and outside service providers. Third, \emph{accountability}: the institution loses control when affected people cannot understand, question or correct a consequential conclusion about them. Evidence exists at each level: proctoring studies show intrusive monitoring \citep{balash2021}, chatbot logs show sensitive disclosure in processing \citep{mireshghallah2024}, and vendor-policy analysis shows accountability gaps \citep{king2025}. Across these sources, most harms arise from ordinary collection, processing and governance rather than model attacks, which remain hard to execute against deployed systems \citep{duan2024}; SAFE-AI therefore weights gate evidence above model-leakage tests.

Table~\ref{tab:rq1} shows how the levels play out by area. A single support conversation can mix all three areas, so data is classified by content, not by which department bought the tool.

\begin{table}[!t]
\centering\small
\caption{RQ1 risks by area, with example harms and sources.}
\label{tab:rq1}
\begin{tabularx}{\columnwidth}{@{}P{13mm}P{16mm}YP{11mm}@{}}
\toprule
Area & Main risk & Example harm & Source\\
\midrule
Teaching & Purpose & Required proctoring with no real alternative; student privacy concerns about generative AI tools & \cite{balash2021,mutimukwe2026,wang2025}\\
Research & Processing & Sensitive topics missed by name redaction; training-data extraction under strong attacks & \cite{mireshghallah2024,carlini2021,duan2024}\\
Admin & Accountability & Decision a student cannot question or correct; vendor reuse of chat data by default & \cite{cormack2016,king2025}\\
\bottomrule
\end{tabularx}
\end{table}

A useful risk register names a harm pathway, not a vague hazard such as ``AI leakage'': for example, \emph{a case worker with broad retrieval rights sees an unrelated student's counselling notes, which may deter that student from seeking help.} That points straight to a testable control: check the requester's role before any document reaches the model. These three levels become requirements for SAFE-AI: purpose maps to gate G1, processing to G2 and G3, and accountability to G4 and G5.

\subsection[RQ2: Existing assessment approaches]{RQ2: \rqtwo}
Existing approaches already cover much of the problem. The OAIC's AI materials in particular show that a conventional PIA can address AI, and European and Canadian practice treat AI assessment as an extension of data-protection assessment rather than a replacement \citep{rintamaki2026,canadaaia}. Covering a topic is different from improving decisions, and the evaluation literature cannot yet say how much PIAs reduce harm \citep{wairimu2024,iwaya2024}. Table~\ref{tab:comparison} therefore compares contributions, not effectiveness scores. None of these approaches says what to do when evidence is missing, overrides an average for a severe harm, bounds connectors and retrieval, or tracks summaries, embeddings and logs after deletion.

\begin{table*}[!t]
\centering\small
\caption{Existing approaches and the deployment-specific work still required. This is an interpretive comparison, not an institutional audit.}
\label{tab:comparison}
\begin{tabularx}{\textwidth}{@{}P{32mm}YY@{}}
\toprule
Approach & Contribution retained & Work made explicit in SAFE-AI\\
\midrule
OAIC PIA guide & Project context, data flows, consultation and risk response. & Versioned AI configuration and linked evidence for each harm scenario.\\
OAIC AI guidance / checklist & AI supplier, input, output and training-use questions. & Required evidence, a responsible reviewer and the consequence of an unanswered question.\\
Educational analytics frameworks & Learner context, participation and appropriate intervention. & Prompts, model evidence, connectors and inventories of derived data.\\
GDPR DPIA / EU AI Act FRIA & Mandatory assessment for high-risk processing, including education; reuse of data-protection work. & Australian jurisdiction, university authority and explicit rules for missing evidence.\\
Canada Algorithmic Impact Assessment & Scored questionnaire mapping answers to impact levels I--IV. & Gates that mitigation cannot offset; severe harm escalated regardless of score.\\
NIST AI RMF / GenAI profile & Lifecycle governance, risk measurement and management. & Local legal scope, university authority and explicit proposed decision thresholds.\\
CSA CCM / AICM & Control catalogues and vendor questionnaires. & A university deployment record; control-level comparison remains to be done.\\
\bottomrule
\end{tabularx}
\end{table*}

\begin{table*}[!t]
\centering\small
\caption{The four decision rules SAFE-AI adds, against what each existing approach specifies (OAIC \cite{oaicpia,oaicchecklist}; NIST \cite{nist2023}; CSA \cite{csaaicm2026}).}
\label{tab:originality}
\begin{tabularx}{\textwidth}{@{}P{28mm}YYY@{}}
\toprule
SAFE-AI rule & OAIC PIA / AI checklist & NIST AI RMF / GenAI & CSA CCM / AICM\\
\midrule
Gates before scores & PIA guide steps 7--8 list risk responses and recommendations; the checklist's five question groups set no stop or hold. & MANAGE~1.1 asks whether deployment should proceed; risk tolerance is set by each organisation (GOVERN~1.3). & AICM controls and CAIQ questionnaire; no decision order.\\
Unknown is not low risk & No status for an unanswered question; follow-up is left to the assessor. & Risks that cannot be measured are documented (MEASURE~1.1); no hold rule. & Unanswered CAIQ items need follow-up; no decision route.\\
Supported with a named Action & Recommendations in the report (step 8); not tied to a stated use boundary. & Responses (mitigate, transfer, avoid, accept) are documented (MANAGE~1.3). & Remediation through control owners; not tied to a stated use boundary.\\
Severity override ($I{=}4$) & Likelihood and consequence judged case by case. & Likelihood and magnitude of impact identified (MAP~5.1); no fixed override. & Severity is framed by control, not harm to people.\\
\bottomrule
\end{tabularx}
\end{table*}

What SAFE-AI adds is the four decision rules in Table~\ref{tab:originality}. The comparative test supports this: assessors using the full OAIC method independently reported that it had no prompt for where AI inference runs, vendor changes after approval, less intrusive alternatives or who receives outputs (Section~\ref{sec:panel}).

\section{The SAFE-AI Framework}
\label{sec:framework}
SAFE-AI is a procedure a university follows before deploying an AI tool and whenever the deployment changes. It does not replace OAIC, NIST or industry catalogues; it adds the decision rules in Table~\ref{tab:originality}. It has four parts: a twenty-criterion checklist grouped under five gates (Appendix~\ref{app:checklist}), a register of harm scenarios, a register of evidence and a signed decision record (Appendix~\ref{app:record}).

\subsection{The assessment record}
Each assessment has an identifier and version and describes exactly one deployment configuration (Section~\ref{sec:context}): purpose, sponsor, affected people, legal setting, data types, a data-flow diagram, product tier, model version and enabled features, plus a less intrusive alternative. Each data flow records sender, recipient, purpose, location, retention and access, so that one vendor-wide privacy statement cannot stand in for a specific setup.

Each criterion receives one of five statuses, used consistently throughout this paper:
\begin{itemize}
\item \emph{Supported}: a reviewer has accepted evidence for the stated claim;
\item \emph{Supported with Action}: the evidence is adequate for a clearly stated use, and a named person must close a recorded gap by a deadline before personal data is used (test below);
\item \emph{Unknown}: the criterion applies, but evidence that could show it is met has not been provided;
\item \emph{Not applicable}: the reviewer records in one line why the criterion's subject is absent from this configuration (for example, no research data, so \crit{06} does not apply);
\item \emph{Failed}: the evidence, \emph{including the proposal's own design}, shows the requirement is not met.
\end{itemize}
The line between the last two matters most, so SAFE-AI uses one rule: \emph{silence is Unknown, contradiction is Failed}. A consent form that says nothing about AI processing leaves \crit{06} Unknown until the ethics documents are checked; a consent form that promises in-house transcription while the design sends recordings to an outside AI service makes \crit{06} Failed. Likewise, a notice promising aggregate reporting fails \crit{03} if the design reports on groups small enough to identify people. Where a safeguard is Unknown, the residual harm rating stays equal to the inherent rating. A residual rating may assume the safeguard of an open blocking Action, because use cannot start until it is verified; if verification fails, the rating reverts and the route is recomputed.

Evidence must match the claim it supports: a published policy proves only what was published, a contract proves an obligation but not that it is honoured, and an encryption certificate says nothing about where data is processed. Table~\ref{tab:evidence} sets out what counts. Each entry records its type, scope, date and reviewer, and expires when the configuration changes.

\begin{table}[!t]
\centering\small
\caption{What counts as evidence in SAFE-AI.}
\label{tab:evidence}
\begin{tabularx}{\columnwidth}{@{}P{14mm}YP{19mm}@{}}
\toprule
Strength & Examples & Supports\\
\midrule
Strong & Independent test or audit of this configuration; configuration export from the university's tenant; signed contract clause & Supported\\
Moderate & Vendor's published, product-specific policy; certification covering this service & Supported for that claim, or with an Action\\
Weak & Marketing pages; general privacy statements; certification for another product & Unknown\\
None & Verbal assurance; evidence for a different version or setup & Unknown\\
\bottomrule
\end{tabularx}
\end{table}

\subsection{Five mandatory gates}
\begin{table}[!t]
\centering\small
\caption{The twenty SAFE-AI criteria, grouped by gate. Appendix~\ref{app:checklist} gives the evidence expected and the source of each.}
\label{tab:criteria}
\begin{tabularx}{\columnwidth}{@{}P{8mm}Y@{}}
\toprule
ID & The assessor asks\ldots\\
\midrule
\multicolumn{2}{@{}l}{\textit{G1 Purpose and permission}}\\
\crit{01} & Is the purpose necessary, with a less intrusive alternative considered?\\
\crit{02} & Which laws, rules and permissions govern this use?\\
\crit{06} & Do research permissions cover this AI processing?\\
\multicolumn{2}{@{}l}{\textit{G2 Processing boundary}}\\
\crit{04} & Are all processing stages and recipients mapped?\\
\crit{07} & Are training and reuse rights separately controlled?\\
\crit{08} & Where do storage, processing and subprocessors operate?\\
\crit{15} & Are supplier claims specific, enforceable and monitored?\\
\multicolumn{2}{@{}l}{\textit{G3 Technical safeguards}}\\
\crit{05} & Is collection limited, including sensitive or inferred information?\\
\crit{09} & What persists, and how are deletion and withdrawal handled?\\
\crit{10} & Is re-identification risk assessed in context?\\
\crit{11} & Is model-leakage evidence relevant to this deployment?\\
\crit{12} & Can prompts, retrieval or tools cross an authorised boundary?\\
\crit{13} & Can outputs disclose or create wrong personal information?\\
\crit{16} & Are access and storage controls appropriate?\\
\multicolumn{2}{@{}l}{\textit{G4 People and recourse}}\\
\crit{03} & Do affected people understand, and have real choice?\\
\crit{14} & Can a person obtain review, correction and challenge?\\
\multicolumn{2}{@{}l}{\textit{G5 Accountable operation}}\\
\crit{17} & Who proposes, reviews, approves and operates it?\\
\crit{18} & Is auditing achieved without excessive logging?\\
\crit{19} & Can incidents be detected, contained and remedied?\\
\crit{20} & What changes or expiry events force reassessment?\\
\bottomrule
\end{tabularx}
\end{table}

Before any risk scores are considered, five gates must be passed (Table~\ref{tab:criteria}):
\begin{itemize}
\item \textbf{G1 Purpose and permission}: the use is justified and the legal, institutional and research permissions cover it;
\item \textbf{G2 Processing boundary}: it is known where data goes, who receives it and what suppliers have committed to;
\item \textbf{G3 Technical safeguards}: the safeguards needed for each disclosure path are in place;
\item \textbf{G4 People and recourse}: affected people have notice, real choice where required, human review and a way to challenge;
\item \textbf{G5 Accountable operation}: there are named owners, an incident response and reassessment triggers.
\end{itemize}
Each gate is \emph{Pass}, \emph{Fail} or \emph{Unknown}. A gate fails if any of its criteria is Failed, and is Unknown if any is Unknown. A failed gate means \emph{stop and redesign}; an unknown gate means \emph{hold for evidence}. A research dataset whose consent excludes or contradicts the proposed AI processing fails G1, however much time the tool would save. Reviewers, not sponsors, decide when a criterion is Not applicable, so a difficult question cannot be dismissed because a supplier will not answer it.

\emph{Supported with Action} is the one controlled exception to ``unknown means hold'', and it has a two-part test. First, the existing evidence must already establish the facts that define the boundary: \emph{where} data goes, \emph{who} may access it, including supplier and support staff, \emph{whether} the use is legally and ethically permitted, and \emph{how long} data is kept or how its retention is controlled. Second, the remaining gap must be something the university can close itself, such as setting a retention schedule or writing study-level guidance, or must only make an already-published commitment enforceable, such as a signed contract clause. A missing ethics approval, an undisclosed processing location or a vendor that will not state its retention therefore stays Unknown. Every Action is \emph{blocking}: it has an owner (a named person or accountable role) and a deadline before personal-data use, and it sets $A$ in the decision rule below. Advice that does not need to be completed before use is recorded separately as a non-blocking recommendation.

Appendix~\ref{app:rules} gives assessor rules for borderline cases.

\subsection{Rating harms and reaching a decision}
For each harm scenario $s$, the assessor rates likelihood $L_s$ and impact $I_s$ on a four-point scale, rating privacy risk by likelihood and severity as in the CNIL PIA method \cite{cnilpia}. Likelihood: 1, remote under documented conditions; 2, plausible but constrained; 3, credible in ordinary foreseeable use; 4, expected or recurring without further controls. Impact: 1, minor and easily reversed, such as information that is already public or trivial; 2, meaningful but limited and remediable, such as ordinary personal details seen by someone who should not see them; 3, serious exposure or a consequential lost opportunity; 4, severe, lasting or very hard to remedy. Assessors explain each rating in terms of duration, sensitivity, scale, vulnerability and reversibility. The register covers privacy harms only; other harms, such as accessibility or academic integrity problems, are referred to the relevant institutional process. A short \emph{cumulative-harm review} then records whether several moderate harms affecting the same people combine into a more serious one; if they do, that combination is rated as its own scenario.

The priority index is $R_s=L_s I_s$. These are ordered labels, not measured quantities, so multiplying them is a transparent sorting convention, not an estimate of expected loss; risk matrices built this way should be used with caution \citep{cox2008}. For that reason SAFE-AI never averages scores, which would let many small concerns dilute one serious one. It keeps the largest residual index, $R_{\max}=\max_s R_s$, shows every individual rating, and adds a \emph{severity override}: any residual impact of 4 is escalated even if its likelihood is 1, unless a gate has already stopped or held the case.

With $I_s$ the residual impact, let $F$ mean any gate has failed; $U$ mean the scope, a gate, an evidence entry, a harm record or the cumulative-harm review is incomplete; $E$ mean a reviewer has requested extra escalation, with a recorded reason; and $A$ mean an open blocking Action or incomplete Action record. The decision rule is:
\begin{equation}
\label{eq:route}
\footnotesize
D=\begin{cases}
\text{Stop and redesign}, & F;\\
\text{Hold for evidence}, & \neg F\land U;\\
\text{Escalate}, & \neg F\land\neg U\land (\max_s I_s{=}4 \lor R_{\max}{\geq}9 \lor E);\\
\text{Conditional review}, & \text{else if }(R_{\max}{\geq}4 \lor A);\\
\text{Standard review}, & \text{otherwise.}
\end{cases}
\end{equation}
Read as five ordered questions, the first ``yes'' decides the route:
\begin{enumerate}[leftmargin=*,itemsep=1pt,topsep=2pt]
\item Has any gate failed? Then \emph{stop and redesign}.
\item Is anything still unknown or incomplete? Then \emph{hold for evidence}.
\item Is any harm severe ($I{=}4$), any harm scored 9 or more, or has a reviewer asked to escalate? Then \emph{escalate}.
\item Is any harm scored 4 or more, or is an Action still open? Then \emph{conditional review}.
\item Otherwise, \emph{standard review}.
\end{enumerate}

\emph{Standard review} means the case may follow the ordinary approval path; \emph{Conditional review} requires written conditions with an owner, deadline and verification before personal data is used; \emph{Escalate} requires a relevant specialist and cannot override a failed gate; any open Actions must still be closed. The thresholds are explicit but uncalibrated, and each institution must test and agree its own. The rule recommends a route; a named person still records the decision.

\subsection{Controls for the three anomalies}
A no-training promise cannot fix all three anomalies; each control must target its own pathway. Table~\ref{tab:anomaly} gives, for each anomaly, a concrete technique and the test that shows it works. Tests use authorised environments and artificial data; they are proposed, not performed here, and where one is infeasible the permitted use must narrow. Table~\ref{tab:traceability} in Appendix~\ref{app:checklist} links each control back to its source literature.

\begin{table*}[!t]
\centering\small
\caption{Controls for each anomaly: technique, the test that shows it works, and the route reached in Section~\ref{sec:eval}.}
\label{tab:anomaly}
\begin{tabularx}{\textwidth}{@{}P{9mm}P{17mm}YYP{30mm}@{}}
\toprule
Anomaly & Criteria, gate & Technique & Pass test & Route in Section~\ref{sec:eval}\\
\midrule
A1 & \crit{01}, \crit{03}, \crit{14}; G1, G4 & Non-AI alternative with the same deadline and marks; human review of any automated flag & Alternative documented in the unit outline; a test flag is reviewed and answered by a named person within the set time & ChatGPT Edu: Conditional, with an Action for a non-Edu pathway\\
A2 & \crit{05}, \crit{06}, \crit{10}; G1, G3 & Redact names and sensitive topics before processing; suppress any report on a group smaller than a set minimum; delete summaries, embeddings and backups with the source & Synthetic transcripts show no residual identifiers; small-group reports are refused; a withdrawal request removes every derived copy & Lyrebird: Conditional with the ethics pack; Hold without it\\
A3 & \crit{04}, \crit{08}, \crit{12}, \crit{15}; G2, G3 & Separate storage and processing map; approved-connector list; role check before any document reaches the model; locations stated in the contract & A test user cannot retrieve another person's record; a hidden instruction in a document (prompt injection) cannot trigger disclosure; an unlisted connector is blocked & ChatGPT Edu: Conditional; Escalate if case files raise impact to 4\\
\bottomrule
\end{tabularx}
\end{table*}

\subsection{Implementation and integration}
\begin{figure*}[!t]
\centering
\includegraphics[width=\textwidth]{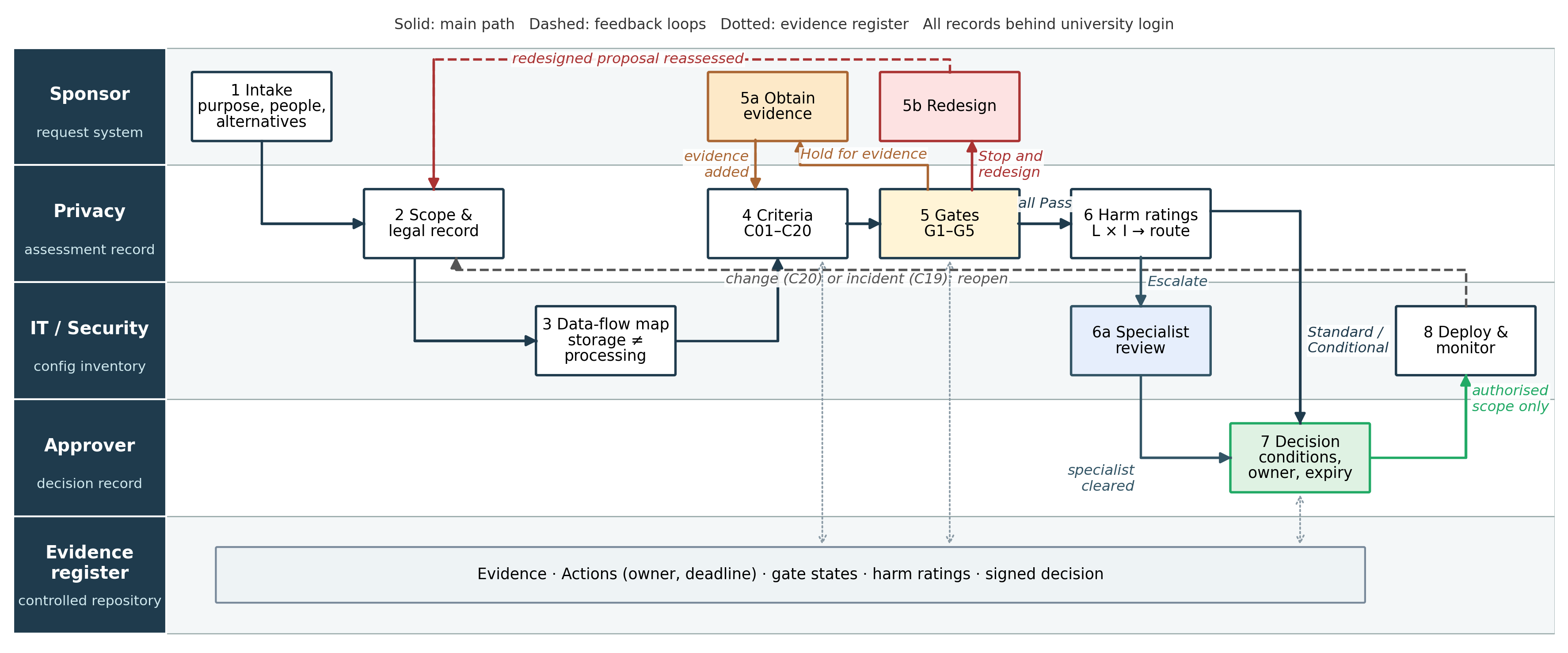}
\caption{SAFE-AI implementation workflow as swimlanes for the sponsor, Privacy, IT/Security and the approver, above a shared evidence register. Steps 1--8 form the main path. At step~5, a failed gate returns the proposal for redesign (5b) and an unknown gate holds it while evidence is obtained (5a); at step~6, a severe or high-priority harm goes to specialist review (6a); a change trigger or incident reopens the assessment. The routing rule never replaces institutional authority.}
\label{fig:workflow}
\end{figure*}

Figure~\ref{fig:workflow} shows the workflow as swimlanes, one per role. The sponsor opens a request in the university's existing request or procurement system; Privacy sets the scope and keeps the assessment record in an access-controlled document store; IT confirms configuration boundaries against its configuration inventory; and an approver independent of the proposer, where practicable, signs the decision. All records sit behind the university login (single sign-on), so access can be limited and logged.

Reviewer capacity, not checklist length, limits scale, so SAFE-AI is applied in two tiers. A \emph{light check} records scope and the five gates only, and suits tools that process no personal information or only public data, such as the clean test dossier. Anything that touches sensitive information, adds a connector or changes processing location receives the \emph{full assessment}. Effort per tier was not measured here and should be timed when SAFE-AI is first trialled with human assessors. Rollout runs in three stages: synthetic cases alongside the existing PIA, shadow assessments with no automated approval, then decision support.

\subsection[RQ3: Additional assessment criteria]{Answer to RQ3: \rqthree}
Compared with existing guidance, the additional criteria concern \emph{specificity} and \emph{evidence}. Assessors need the exact account and feature configuration; separate permissions for training, inference and retrieval; sensitivity judged in context, not just by direct identifiers; model-leakage evidence under stated assumptions; and the fate of outputs, embeddings, logs and backups. University-specific additions cover weak choice in teaching, research permissions and challenge routes, and every safeguard has a named owner.

Table~\ref{tab:law} maps the criteria to the privacy principles that apply to Australian universities: the Australian Privacy Principles (APPs) for ANU and private universities, and the NSW Information Protection Principles (IPPs) for NSW public universities \citep{oaicapp,nswppip}. Criteria \crit{10}--\crit{12} and \crit{20} have no direct counterpart: they cover AI-specific risks that current principles do not name, which is part of what SAFE-AI adds.

\begin{table}[!t]
\centering\small
\caption{Criteria mapped to Australian privacy principles (indicative; local legal advice still applies).}
\label{tab:law}
\begin{tabularx}{\columnwidth}{@{}YP{12mm}P{17mm}@{}}
\toprule
Criteria & APP & NSW IPP\\
\midrule
Necessity, minimisation (\crit{01}, \crit{05}) & 3 & 1, 4\\
Notice and choice (\crit{03}) & 5 & 3, 6\\
Use, reuse, training (\crit{06}, \crit{07}) & 6 & 10, 11\\
Overseas processing (\crit{08}, \crit{15}) & 8 & 12 (s~19)\\
Output accuracy (\crit{13}) & 10 & 9\\
Retention, security, audit (\crit{09}, \crit{16}, \crit{18}) & 11 & 5\\
Access, correction (\crit{14}) & 12, 13 & 7, 8\\
Governance, mapping (\crit{02}, \crit{04}, \crit{17}, \crit{19}) & 1; NDB & MNDB\\
Health information (\crit{05}, \crit{08}) & 3, 8 & HPPs incl.~14 (HRIP Act)\\
No direct counterpart (\crit{10}--\crit{12}, \crit{20}) & -- & --\\
\bottomrule
\end{tabularx}
\end{table}

\section{Evaluation and Discussion}
\label{sec:eval}
We chose two public university deployments that between them illustrate all three anomalies. Both assessments use only published pages, checked on 21~September and 10~October~2026. We had no access to contracts, ethics files or internal systems, so each result is our reading of public text, not an assurance finding by the university concerned. Each case asks one decision question and ends with the route produced by Equation~\eqref{eq:route}.

\subsection{Worked example 1: ANU Project Lyrebird}
Project Lyrebird is ANU's AI-assisted research-interview platform, and it illustrates anomaly A2. The decision question is: \emph{could a privacy reviewer approve Lyrebird for a new interview study that may collect sensitive personal stories, using only its published privacy policy} \citep{anulyrebird}? Table~\ref{tab:lyrebird} records the decisive public claims and the status each supports.

\begin{table*}[!t]
\centering\small
\caption{SAFE-AI applied to Project Lyrebird using the public privacy policy only.}
\label{tab:lyrebird}
\begin{tabularx}{\textwidth}{@{}P{18mm}P{24mm}YP{32mm}@{}}
\toprule
Criterion & Status & What the page supports & Reviewer implication\\
\midrule
\crit{04}, \crit{08} & Supported & Storage at NCI, Canberra; AI processing on AWS Bedrock in Sydney; Cloudflare sees connection data only & Processing map complete.\\
\crit{01}, \crit{06} & Unknown until ethics pack attached & Study consent and ethics approval take precedence over the platform policy & Supported once the study's ethics pack is attached; until then G1 is Unknown.\\
\crit{02} & Supported & Privacy Act rights and ANU privacy policy cited & Adequate for ordinary operation.\\
\crit{07} & Supported & AWS processes text to reply; stated not retained or used for training & Accepted on the published claim.\\
\crit{09}, \crit{14} & Supported & Immediate deletion on withdrawal; backups kept at most 7~days; withdrawal link & Routes described.\\
\crit{05}, \crit{10} & Supported with Action & No name or IP address in the research record; optional email for withdrawal only & Study-level minimisation still required.\\
\crit{12} & Not applicable & No connectors or tools beyond the published Australian interview path & Becomes applicable if connectors are enabled.\\
\crit{13} & Supported with Action & Only researchers access study content & Output-use limits needed in the information sheet.\\
\crit{15} & Supported with Action & Named processors (AWS Bedrock, NCI, Cloudflare); region and no-retention claims; no contract seen & Contract extract still required.\\
\crit{17}, \crit{19} & Supported & ANU as controller; named HDRH and Privacy Officer contacts; breach-notification intent & Owners named.\\
\crit{18}, \crit{20} & Supported with Action & Restricted administrator access; notice of material policy changes & Audit record and change triggers needed.\\
\bottomrule
\end{tabularx}
\end{table*}

The three criteria not in the table are \crit{03}, Supported with Action: participants may decline, stop or withdraw, but study-specific notice is required; \crit{11}, Not applicable because no local training is proposed; and \crit{16}, Supported by the stated restriction of administrator access.

\noindent\textbf{Route.} On the public policy alone, \crit{01} and \crit{06} are Unknown, so G1 is Unknown and the route is \emph{Hold for evidence}. With the study ethics pack attached, all five gates Pass. The gaps in \crit{03}, \crit{05}, \crit{10}, \crit{13}, \crit{15}, \crit{18} and \crit{20} meet the two-part test, because the policy already establishes where data goes, who may access it, the permission route and retention, and each gap is closable by ANU or adds enforceability; each is owned and due before the first participant ($A{=}1$, $F{=}U{=}0$). Two sample harms: (i)~a distinctive story in a small study identifies a participant in published quotes, residual $L{=}2$, $I{=}3$ ($R{=}6$) after study-level minimisation; (ii)~a study's information sheet omits the platform's withdrawal route, $L{=}3$, $I{=}2$ ($R{=}6$). The cumulative-harm review found no combined harm, so with $R_{\max}{=}6$ the route becomes \emph{Conditional review}.

\subsection{Worked example 2: University of Sydney ChatGPT Edu}
ChatGPT Edu is the University of Sydney's institutional workspace for general learning use, and it illustrates anomaly A3, with A1 arising because a campus-wide tool weakens real choice \cite{sydneyedu,sydneyedunews,sydneyuse}. The decision question is: \emph{would SAFE-AI support enabling the service for ordinary learning assistance, using only published pages?} The service page states that data is stored in Australia at rest but that prompts may be processed outside Australia, which is exactly the split described by A3. It schedules a student pilot for October~2026 and release to all students in Semester~1~2027. Table~\ref{tab:sydneyedu} records the decisive claims.

\begin{table*}[!t]
\centering\small
\caption{SAFE-AI applied to University of Sydney ChatGPT Edu using public student and news pages only.}
\label{tab:sydneyedu}
\begin{tabularx}{\textwidth}{@{}P{18mm}P{24mm}YP{30mm}@{}}
\toprule
Criterion & Status & What the pages support & Reviewer implication\\
\midrule
\crit{04}, \crit{08} & Supported & Australian data residency at rest; prompts may be processed transiently outside Australia, in the countries OpenAI lists for its Edu infrastructure subprocessors (e.g. United States, Ireland, Japan) \cite{openaisub} & Two separate paths: storage in Australia, processing overseas (A3).\\
\crit{02} & Supported & The privacy notice names the PPIP and HRIP Acts and states personal information in AI tools may be processed outside Australia \cite{sydneynotice} & Adequate for general learning.\\
\crit{07} & Supported & University data not used to train vendor models & Reopen for high-sensitivity use.\\
\crit{16}, \crit{18} & Supported & Identity management, encryption, audit logging & Described at enterprise level.\\
\crit{09} & Supported with Action & Configurable retention; administrator access for security and governance & Local retention schedule and admin-access purpose needed.\\
\crit{03}, \crit{14} & Supported with Action & Responsible-use guidance; advice to avoid unnecessary personal data & Choice weak if expected for coursework (A1).\\
\crit{05}, \crit{13} & Supported with Action & Guidance against unnecessary personal inputs; AI assists but does not replace judgement & Enforced guardrails needed; students may still paste personal details.\\
\crit{15} & Supported with Action & OpenAI staff access limited to incidents \cite{openaient}; subprocessors listed; no contract seen & An enforceable contract extract is still required.\\
\crit{19} & Supported with Action & University incident processes; supplier incident commitments \cite{openaient}; no service-level incident owner named & A named service incident owner is still required.\\
\crit{12}, \crit{20} & Supported & Higher-risk features disabled until separately assessed; staged rollout & Enabling connectors or agents reopens G2 and G3.\\
\bottomrule
\end{tabularx}
\end{table*}

Of the criteria not in the table, \crit{01} and \crit{17} are Supported for general learning; \crit{06}, \crit{10} and \crit{11} are Not applicable because no research transcripts, linked or released dataset, or local model training are in scope.

\noindent\textbf{Route.} All five gates Pass for general learning. The gaps in \crit{03}/\crit{14}, \crit{05}/\crit{13}, \crit{09}, \crit{15} and \crit{19} pass the two-part test, because the pages establish storage and processing locations, administrator access and configurable retention; they are blocking Actions owned by faculties, ICT and procurement ($A{=}1$, $F{=}U{=}0$). Three sample harms (residual): (i)~a student pastes names and marks into overseas processing, $L{=}3$, $I{=}2$ ($R{=}6$) after enforced input guardrails; (ii)~administrator-accessible logs keep prompts beyond purpose, $L{=}2$, $I{=}3$ ($R{=}6$) after role limits; (iii)~anomaly A1, a student cannot practically opt out once the tool is expected for group work, $L{=}3$, $I{=}2$ ($R{=}6$) with a documented alternative. A compulsory tool with no alternative would instead be Failed. The cumulative-harm review found no combined harm. With $R_{\max}{=}6$ and $\max I_s{=}3$, the route is \emph{Conditional review}; counselling or misconduct content could raise impact to 4 and the route to \emph{Escalate}.

\subsection{Comparative test with realistic dossiers}
\label{sec:panel}
\noindent\textbf{Design.} The worked examples cannot measure accuracy, because their correct answer is unknown. We therefore wrote eight fictional dossiers, each holding the four to six documents a reviewer actually receives (request form, contract or terms extract, security questionnaire, notices, configuration exports, emails), with problems visible only by comparing documents (Appendix~\ref{app:eval}). Six contain a blocking problem, covering all three anomalies; one is clean; one should be approved with conditions. Expected routes were fixed before assessment. Five simulated assessors all used one small model (Claude Haiku), so that differences came from realistic personas rather than model capability: a legalistic privacy officer, a time-pressed security architect who trusts certifications, a research ethics manager, a sceptical student caseworker and a new procurement officer who relies on vendor statements. Each assessed every dossier with SAFE-AI (Form~A) and, in a separate fresh session, with a full OAIC-based form covering the guide's ten steps, an APP~1--13 check and all five AI-checklist question groups (Form~B).

\begin{table}[!t]
\centering\small
\caption{Simulated panel results (5 personas $\times$ 8 dossiers per method). $\kappa$ is computed as in Appendix~\ref{app:eval}.}
\label{tab:panel}
\begin{tabularx}{\columnwidth}{@{}Ycc@{}}
\toprule
Measure & SAFE-AI & Full OAIC\\
\midrule
Blocking problems caught (/30) & 30 & 28\\
Conditional case correct (/5) & 4 & 4\\
Clean case approved (/5) & 0 & 1\\
Clean case blocked (/5) & 1 & 0\\
Fleiss' $\kappa$ (route agreement, $N{=}8$ dossiers) & 0.81 & 0.48\\
\bottomrule
\end{tabularx}
\end{table}

\noindent\textbf{Results.} SAFE-AI caught every blocking problem (Table~\ref{tab:panel}); its routes matched the expected route or an accepted alternative in 34 of 40 assessments. The full OAIC form missed two: the security persona approved compulsory keystroke and writing-style profiling with conditions, and the ethics persona approved automatic voiceprinting added to an approved captioning tool, which all five SAFE-AI assessors caught through the reassessment trigger (\crit{20}). Under the proportionate tier rule, four of five SAFE-AI assessors gave the low-risk tool Conditional review, recording its missing alternatives, triggers and incident contact as Actions; OAIC matched this. Neither method reliably approved the clean case: most assessors attached conditions, and one SAFE-AI assessor held it. SAFE-AI assessors agreed more ($\kappa$ 0.81 versus 0.48).

\noindent\textbf{Interpretation.} SAFE-AI's gates stopped every blocker even for assessors inclined to trust vendors, and its explicit rules for triggers and human review caught problems the OAIC form let through. The proportionate tier let the low-risk tool proceed with conditions, although the clean case still attracted unnecessary conditions under both methods. With five assessors and eight dossiers, the agreement gain is indicative only.

\subsection[RQ4: Supporting acceptability decisions]{Answer to RQ4: \rqfour}
Whether a risk is acceptable remains a human institutional judgment. SAFE-AI supports that judgment in four ways visible in the two cases: it stops known failures, holds cases when essential evidence is missing, escalates severe harms regardless of an overall score, and records the conditions, owners, deadlines and use boundary behind each decision so it can be reviewed later. Benefits never cancel out unlawful processing or an unresolved severe harm; instead they enter the earlier question of whether the same goal could be met with less intrusive processing.

\subsection{Limitations}
SAFE-AI never approved the clean test dossier: attaching conditions to safe tools adds paperwork for sponsors and reviewers, and that workload was not measured. The authors wrote both the dossiers and their expected routes, and one model family wrote the dossiers and acted as every assessor, so assessors may share the writers' reading of the documents. The authors also designed SAFE-AI and applied it to both public cases, so those routes show only that the rules can be applied as written, not that independent assessors would reach the same routes. Others might reasonably disagree with some of our Not-applicable and Supported-with-Action calls.

\section{Recommendations}
\textbf{University leadership} should approve each defined deployment, naming its purpose, population and enabled features, rather than a product for every use, and reassess when scope changes. The trade-off is slower rollout; the cases show why: one ChatGPT Edu approval would otherwise silently cover counselling content.

\textbf{Privacy and research-governance teams} should pilot SAFE-AI as an extension of existing review and agree who owns each gate. Gates add reviewer workload, so the light-check tier should be used for public-data tools.

\textbf{Procurement and IT} should ask for evidence claim by claim (Table~\ref{tab:evidence}) and reopen the assessment when training, location, access, retention or connectors change. A Hold can delay deployment while a vendor responds, so evidence requests should carry deadlines (Appendix~\ref{app:rules}).

Before relying on the thresholds, universities should pilot SAFE-AI alongside their current PIA with human assessors and calibrate them locally.

\section{Conclusion}
Universities are deploying AI tools faster than they can show what those tools do with personal information. The principles already exist in Australian guidance and university policy. What is missing is a way to apply them to one deployment at a time, with evidence, and with a named person accountable for the decision.

SAFE-AI provides that step through twenty criteria, five mandatory gates, a register of harm scenarios and a decision rule in which missing evidence pauses a decision and severe harm can never reach standard review. Applied to ANU's Project Lyrebird and the University of Sydney's ChatGPT Edu using public pages alone, it held Lyrebird until each study's ethics approval is attached, after which it reaches Conditional review; gave ChatGPT Edu Conditional review with named Actions; and separated ChatGPT Edu's Australian storage from its possible overseas processing. In a comparative test with persona-based AI assessors on realistic dossiers it caught every blocking problem, two more than a full OAIC-based assessment, while matching it on a low-risk conditional case, although it never fully approved a clean one.

In summary, Section~\ref{sec:findings} answered RQ1 (risks at purpose, processing and accountability levels) and RQ2 (existing approaches lack decision rules); Section~\ref{sec:framework} answered RQ3 (configuration-specific, evidence-backed criteria); and Section~\ref{sec:eval} answered RQ4 (gates, holds, escalation and recorded conditions). Its limits are that both cases used public pages only, the comparative test used simulated assessors, and the thresholds are uncalibrated. Future work should trial SAFE-AI with human privacy and IT staff and calibrate thresholds locally. For the student in our opening example, the value is simple: someone has checked what is known, what is missing and who is responsible before their information enters the system.

\section*{Acknowledgment}
The authors thank their tutor and unit teaching staff. AI-assisted drafting is disclosed in Appendix~\ref{app:ai}. All authors remain responsible for verifying claims and sources.

\bibliographystyle{IEEEtran}
\bibliography{references}

\clearpage
\appendices
\onecolumn
\section{SAFE-AI assessment checklist}\label{app:checklist}
Give every criterion one of the five statuses in Section~\ref{sec:framework}: Supported, Supported with Action, Unknown, Not applicable (with a reviewer-approved reason) or Failed. Record the evidence ID, version, date, scope, reviewer, control owner and any outstanding Action. ``Yes'' without evidence is Unknown. The criteria are a proposed synthesis; the source column shows where each came from, not external validation of the checklist. A ``complete'' record means the required fields are present and current; it does not prove the underlying claim.

\begingroup\small
\setlength{\baselineskip}{11pt}
\setlength{\tabcolsep}{4pt}
\begin{longtable}{@{}P{9mm}P{52mm}P{57mm}P{33mm}@{}}
\caption{Twenty assessment criteria, evidence expectations and sources.}\label{tab:checklist}\\
\toprule
ID & Assessment question & Minimum relevant evidence / action & Derivation\\\midrule\endfirsthead
\multicolumn{4}{l}{\textit{Table \ref{tab:checklist}, continued}}\\\toprule
ID & Assessment question & Minimum relevant evidence / action & Derivation\\\midrule\endhead
\bottomrule\endfoot
C01 & Is the specific purpose necessary and proportionate, with a less intrusive alternative considered? & Purpose statement, benefits, alternatives and reasons for rejecting them; do not use convenience as the only justification. & Slade and Prinsloo \cite{slade2013}; Pardo and Siemens \cite{pardo2014}; OAIC PIA \cite{oaicpia}\\[4pt]
C02 & Which laws, institutional rules and permissions govern this use? & Jurisdiction and obligations record; relevant authorisation and institutional interpretation; research approval where required. & OAIC scope \cite{oaicscope}; University of Sydney notice \cite{sydneynotice}; Monash policy \cite{monashpolicy}\\[4pt]
C03 & Who is affected, what do they understand, and can they exercise meaningful choice where required? & Population and vulnerability analysis, consultation record, accessible notice and practical alternative where applicable. & DELICATE \cite{drachsler2016}; proctoring studies \cite{balash2021,mutimukwe2026}\\[4pt]
C04 & Are all processing stages and recipients mapped for this configuration? & Versioned diagram covering collection, inference, retrieval, training, outputs, logs and connected services. & OAIC AI checklist \cite{oaicchecklist}; University of Sydney Edu \cite{sydneyedu}\\[4pt]
C05 & Is collection limited, including sensitive meaning and inferred information? & Field and narrative sensitivity review, minimisation choices and tests appropriate to the proposed data. & Liu and Khalil \cite{liu2023}; Mireshghallah et al. \cite{mireshghallah2024}\\[4pt]
C06 & Do research permissions cover this AI use and secondary processing? & Study-specific consent / information sheet and ethics documentation; record any required amendment. & OAIC training guidance \cite{oaictrain}; ANU Lyrebird \cite{anulyrebird}\\[4pt]
C07 & Are training, adaptation and other reuse rights separately controlled? & Product-tier contract, tenant settings and a documented decision on input/output reuse; distinguish commitments from enforcement. & King et al. \cite{king2025}; OAIC AI guidance \cite{oaicuse}\\[4pt]
C08 & Where do storage, inference, support and subprocessors operate? & Processing-region and recipient schedule, remote-access arrangements and jurisdiction review; separate location types. & University of Sydney Edu \cite{sydneyedu}; OAIC checklist \cite{oaicchecklist}\\[4pt]
C09 & What persists, and how are deletion, withdrawal and lawful retention reconciled? & Inventory of source records, outputs, embeddings, logs and backups; retention schedule and verified deletion/restoration procedure. & ANU Lyrebird \cite{anulyrebird}; University of Sydney notice \cite{sydneynotice}\\[4pt]
C10 & Is provenance known and re-identification risk assessed in context? & Data-source permissions, linkage threats, contextual review and de-identification limitations; do not equate pseudonyms with anonymity. & LEA in Private \cite{steiner2016}; OAIC training guidance \cite{oaictrain}\\[4pt]
C11 & Is model-related disclosure evidence relevant to the deployment? & Threat model; supplier evidence or authorised tests for applicable inference/extraction threats; limits, attacker access and error rates. & Shokri et al. \cite{shokri2017}; Carlini et al. \cite{carlini2021,carlini2022}; Duan et al. \cite{duan2024}; Hayes et al. \cite{hayes2025}\\[4pt]
C12 & Can prompts, retrieval or tools cross an authorised information boundary? & Access-before-retrieval tests; scoped service credentials; adversarial input tests where appropriate; prohibited connector list. & NIST GenAI profile \cite{nist2024}; proposed technical checks\\[4pt]
C13 & Can outputs disclose or create inaccurate personal information? & Output-use constraints, factual verification, disclosure tests and limits on consequential automated use. & OAIC AI guidance \cite{oaicuse}; Cormack \cite{cormack2016}\\[4pt]
C14 & Can a person obtain review, correction and a meaningful challenge? & Named human reviewer, accessible challenge route, escalation time expectations and correction propagation plan. & Educational analytics ethics \cite{slade2013,pardo2014}; University of Sydney privacy rights \cite{sydneynotice}\\[4pt]
C15 & Are supplier claims specific, enforceable and monitored for change? & Contract version, audit/assurance scope, subcontractor terms, incident notification and change obligations. & Monash policy \cite{monashpolicy}; King et al. \cite{king2025}; OAIC checklist \cite{oaicchecklist}\\[4pt]
C16 & Are access and storage controls appropriate to the data? & Role matrix, authenticated access, encryption/key responsibilities, separation of environments and access-review evidence. & OAIC PIA \cite{oaicpia}; NIST AI RMF \cite{nist2023}\\[4pt]
C17 & Who proposes, reviews, accepts and operates the deployment? & Named role map and approval authority; separation of proposer and reviewer where feasible; recorded acceptance reasons. & NIST AI RMF \cite{nist2023}; Monash policy \cite{monashpolicy}\\[4pt]
C18 & Is auditability achieved without excessive logging? & Minimal audit-event schema, restricted evidence access and a recordkeeping schedule; justify any retained prompt content. & ANU data-governance guidance \cite{anuguidance}; proposed minimisation\\[4pt]
C19 & Can incidents be detected, contained and remedied? & Incident owner, response exercise, suspension control, notification assessment and affected-person remedy pathway. & OAIC PIA \cite{oaicpia}; NIST lifecycle governance \cite{nist2023}\\[4pt]
C20 & What changes or expiry events force reassessment? & Triggers for model, feature, data, purpose, vendor and population changes; named monitoring owner; review date. & NIST AI RMF \cite{nist2023}; University of Sydney feature review \cite{sydneyedu}\\
\end{longtable}
\endgroup

\begin{table}[h]
\centering\small
\caption{Examples of traceability from evidence to decision. Criteria are listed in full in Appendix~\ref{app:checklist}.}
\label{tab:traceability}
\begin{tabularx}{\linewidth}{@{}P{24mm}P{28mm}YY@{}}
\toprule
Evidence & Criteria & Required artefact / control & Decision consequence\\
\midrule
Proctoring studies; educational ethics & \crit{01}, \crit{03}, \crit{14} & Necessity review, alternative pathway and challenge procedure & Missing justification holds; a known unacceptable design fails G1/G4.\\
Conversational disclosure; research guidance & \crit{05}, \crit{06}, \crit{10} & Context-aware redaction and study permission for the exact processing & Unauthorised research reuse fails G1.\\
Model privacy studies & \crit{11} & Threat model, attack conditions and bounded test evidence & An unsupported safety claim stays Unknown.\\
Sydney deployment information; OAIC AI checklist & \crit{04}, \crit{08}, \crit{12}, \crit{15} & Tenant, processing map, supplier terms and feature inventory & An unknown recipient or transfer holds G2.\\
ANU Lyrebird; lifecycle governance & \crit{09}, \crit{18}, \crit{20} & Retention schedule, inventory of derived data and change triggers & Unverified deletion or changed scope reopens review.\\
\bottomrule
\end{tabularx}
\end{table}

\section{Assessor consistency rules}\label{app:rules}
\noindent\textbf{Access and change.} Access criteria (\crit{12}, \crit{16}) are tested against the institution's own authorised access rules; access that is authorised but broader than the purpose needs is not marked Failed, but must be recorded as a harm scenario so that it is rated and, if serious, escalated. When a reassessment trigger under \crit{20} fires, for example a new connector, the approval lapses for the changed part: the criteria it affects return to Unknown and the changed feature is suspended until reassessed. If the change was made without reassessment, that is also recorded as an incident under \crit{19}; the route is then computed for the changed part, while the unchanged deployment continues under its existing approval. If the change, as described, itself shows a requirement is not met, that criterion is Failed; a new capability whose effect is not yet documented is Unknown. The decision record states both routes.

\noindent\textbf{Consistency rules.} (1)~\emph{Proportionality}: a criterion whose subject is present but which this deployment does not change, such as standard incident handling, the institutional privacy notice or an approved platform's security controls, may be marked Supported by citing that existing institutional control as its evidence; this stops routine criteria from holding low-risk cases while still requiring a reference. (2)~\emph{Sharing versus access}: the access rule above covers staff access under institutional rules; disclosing content to people other than those needed to deliver the stated purpose, such as classmates or the public, is judged under \crit{05} and \crit{13} and can be Failed by design. Where access is authorised but broad, likelihood is rated for whether \emph{some} authorised person will see records they do not need; with tens of users and no preventive limit, this is normally 3. (3)~\emph{Safeguard credit}: a safeguard lowers a residual likelihood only if its evidence shows it \emph{prevents} the harm pathway; controls that only detect misuse afterwards, such as audit logs or periodic access reviews, are recorded but do not lower the rating. (4)~\emph{Refusal}: when requesting evidence, the reviewer should set a deadline. Only once that deadline has passed does a supplier's refusal or silence about where data goes, who may access it or how long it is kept make the criterion Failed; before then it remains Unknown. (5)~\emph{Necessity}: if the deployment is compulsory, monitors people or processes sensitive information, \crit{01} is Failed when no less intrusive alternative was considered or the only justification is cost or convenience. (6)~\emph{Proportionate tier}: a deployment is low-risk when it is optional, processes no sensitive information, uses no connector and has an evidenced processing location; for it, a missing administrative item (alternatives analysis, owners, reassessment triggers, incident contact, audit design) is Supported with Action, but the four facts in the two-part test must still be evidenced. Public, non-personal or aggregate information is rated impact~1. (7)~\emph{Meaningful human review}: where staff confirm nearly all automated recommendations about people, with no reasons given and no specific appeal, \crit{14} is Failed. Three clarifications complete the rules. Health, disability, financial-hardship and misconduct information is rated at least impact~3 when seen by someone who does not need it. The necessity rule applies to the justification for the deployment itself, not to the reasons given for rejecting an alternative. An Action with an accountable role and a deadline counts as complete; an incomplete Action record sets $A$, not $U$. The assessor completes the record and the reviewer confirms that the scope, harm register and cumulative-harm review are complete before the route is taken.

\section{Comparative test materials}\label{app:eval}
\noindent\textbf{Dossiers.} The test used eight fictional dossiers set at a fictional NSW public university. A dossier holds four to six realistic documents: the university's project request form (with optional fields for alternatives, reassessment triggers and incident contact), contract, data-processing agreement, terms or order-form extracts with numbered clauses, security questionnaires, notices, procedures or ethics applications, configuration or distribution-list exports, release notes and email threads. Problems were never labelled; each could be found only by comparing documents: (F1)~compulsory keystroke and writing-style profiling of all first-year essays, justified by deterrence and cost; (F2)~research analysis of placement debriefs recorded ``for feedback to you only'', containing patient details and processed in Singapore; (F3)~a CV-matching tool whose inference provider and region are left to a blank order-form section, with no deadline set; (F4)~incident-report summaries, including assault disclosures, emailed to all student resident advisers; (F5)~scholarship screening that rejects hardship applicants automatically with no reasons or appeal; (F6)~a staff IT knowledge search (clean); (F7)~an optional language-practice chatbot, low-risk but missing alternatives, triggers and an incident contact (conditional); (F8)~voiceprint speaker identification switched on by default in an approved captioning tool without the required reassessment.

\noindent\textbf{Expected routes and scoring.} For each dossier the planted problem, criteria, expected SAFE-AI route (with any accepted alternative) and expected OAIC outcome were written and their files hashed before assessment, and not changed afterwards. A blocking problem was caught when the route was Stop, Hold or Escalate (Form~A) or redesign, more information or referral (Form~B). The clean case counted as approved only at Standard review or Approve, and as blocked at a blocking route. The conditional case was correct at Conditional review (Form~A) or Approve or Approve with conditions (Form~B).

\noindent\textbf{Forms.} Form~A records all twenty criteria with quoted evidence, any Actions with owner and deadline, the five gates, up to three harms with likelihood, impact and rationale, the cumulative-harm review, the route and confidence; its guide summarises the statuses, two-part test, gate and route rules and the Appendix~\ref{app:rules} rules. Form~B follows the OAIC guide's ten steps: threshold assessment, scope, project description, stakeholders, information flows, impact analysis, a finding for each of APP~1--13, all five OAIC AI-checklist question groups, risk treatment, recommendations, one of five overall outcomes (approve, approve with conditions, more information needed, refer, redesign) and review triggers. Like the OAIC guide, Form~B gives no rule for choosing the outcome.

\noindent\textbf{Assessors and procedure.} All assessors used the same small model (Claude Haiku), so that differences in judgement came from the persona rather than model capability. Each persona (Table~\ref{tab:roles}) ran a separate fresh session for each form, never saw the other form, the expected routes or another assessor's work, and was told to read every document, use only the dossier, treat anything unstated as not provided, assume the university's standard approved controls unless the dossier said otherwise, quote supporting evidence and not look anything up. SAFE-AI's rules were developed iteratively, including trial assessments on other dossiers; the test dossiers were written afterwards.

\begin{table}[h]
\centering\small
\caption{Assessor personas (identical for both forms).}
\label{tab:roles}
\begin{tabularx}{\linewidth}{@{}P{42mm}Y@{}}
\toprule
Persona & Brief given to the assessor\\
\midrule
Senior privacy officer & 15 years in NSW public-sector privacy; legalistic; accepts no claim without a document; reads contracts clause by clause; holds when evidence is thin.\\
IT security architect & ISO~27001 lead auditor; time-pressed; trusts certifications, encryption and configuration exports; inclined to approve when security is strong and to leave consent questions to others.\\
Research ethics manager & Supports the ethics committee; focused on consent, information sheets and the National Statement; refers when unsure.\\
Student union caseworker & Six years in misconduct and special-consideration cases; sceptical of institutional assurances; focused on power, real choice, appeal routes and vulnerable students.\\
Procurement officer & Three months in role with no privacy training; wants projects to proceed; reads quickly and relies on the guide and vendor statements.\\
\bottomrule
\end{tabularx}
\end{table}

\noindent\textbf{Agreement.} Agreement was computed as Fleiss' $\kappa$ \cite{fleiss1971}, which measures how far several raters agree on a categorical choice beyond chance. Here the raters are the $n{=}5$ assessors, the subjects are the $N{=}8$ dossiers, and the categories are the five SAFE-AI routes or the five OAIC outcomes. With $n_{ij}$ the number of assessors who put dossier $i$ in category $j$:
\begin{equation*}
P_i=\frac{\sum_j n_{ij}^2-n}{n(n-1)},\quad \bar P=\frac{1}{N}\sum_i P_i,\quad p_j=\frac{\sum_i n_{ij}}{Nn},\quad \bar P_e=\sum_j p_j^2,\quad \kappa=\frac{\bar P-\bar P_e}{1-\bar P_e}.
\end{equation*}
$P_i$ is the share of assessor pairs that agree on dossier $i$, $\bar P$ the mean observed agreement and $\bar P_e$ the agreement expected by chance from how often each category was used. For example, with SAFE-AI four assessors chose Conditional review and one Hold for dossier F6, so $P_{F6}=(4^2+1^2-5)/20=0.60$; across all eight dossiers $\bar P=0.900$, $\bar P_e=0.461$ and $\kappa=0.81$. For Form~B, $\bar P=0.637$, $\bar P_e=0.300$ and $\kappa=0.48$. $\kappa{=}1$ means perfect agreement and $\kappa{=}0$ agreement no better than chance; with eight dossiers no confidence interval is reported. The dossiers, expected routes, forms, guides, scoring script and all ten completed records, together with the trial materials, are available from the authors.

\section{Assessment and decision record template}\label{app:record}
\noindent\textbf{Identity.} Assessment ID; version; date; sponsor; system owner; university unit; intended users; affected people; purpose; excluded uses; model/service/version; product tier; tenant; enabled connectors.\\
\textbf{Scope.} Data categories; source permissions; jurisdiction; institutional policy references and versions; research approval reference; intended benefit; less intrusive alternatives; data-flow attachment.\\
\textbf{Evidence.} Criterion ID; claim; evidence reference; documentary/technical/procedural type; scope; date; expiry; status; limitations; reviewer; control owner.\\
\textbf{Harm scenario.} Scenario ID; affected people; actor/event; information; consequence; relevant criteria; inherent rating and rationale; safeguard; residual rating and rationale; uncertainty; test limitations.\\
\textbf{Decision.} Gate states and reasons; recommended route; independent review; final institutional decision; conditions; verification evidence; accountable signatory; permitted data and use; review date; suspension/reassessment triggers.\\
\textbf{Closure.} Date conditions verified; configuration released; outstanding monitoring; affected-person contact; next review. A blank signatory field means no institutional approval has been recorded.

\section{Glossary}\label{app:glossary}
\begingroup\small
\begin{longtable}{@{}P{38mm}P{120mm}@{}}
\toprule
Term & Plain-language meaning\\\midrule\endhead
\bottomrule\endfoot
Agent & An AI feature that can take actions for a user, such as sending email or editing files, not just answer questions.\\
Connector & A link that lets an AI tool read from or write to another system, such as email, calendars or student records.\\
Data residency / storage at rest & Where data is kept when it is not being used. It does not say where data is processed.\\
De-identification & Removing names and other direct identifiers. People may still be recognisable from context.\\
DPIA & Data Protection Impact Assessment: the impact assessment required under the EU's GDPR for high-risk processing.\\
Embedding & A numerical representation of text that AI systems use for search; it can preserve meaning after the original text is deleted.\\
Extraction attack & An attempt to make a model reproduce text it was trained on.\\
FRIA & Fundamental Rights Impact Assessment: the assessment the EU AI Act requires for certain high-risk AI uses.\\
Gate & A mandatory pass/fail check in SAFE-AI that is applied before any risk scores.\\
Inference & An AI model processing a request to produce an answer. It can happen in a different country from storage.\\
Inherent / residual risk & Risk before safeguards / risk remaining after safeguards that are backed by evidence.\\
Membership inference attack & An attempt to work out whether a particular person's data was used to train a model.\\
NDB scheme & Australia's Notifiable Data Breaches scheme, which requires certain breaches to be reported.\\
MNDB scheme & The NSW Mandatory Notification of Data Breach scheme for NSW public agencies, including public universities.\\
APP / IPP & The Australian Privacy Principles (federal) and NSW Information Protection Principles (state).\\
NIST AI RMF & A US government framework for managing AI risk across its lifecycle.\\
OAIC & Office of the Australian Information Commissioner, the federal privacy regulator.\\
Ordinal rating & A ranked label (1 to 4) whose steps are not equal measured amounts.\\
PIA & Privacy Impact Assessment: a structured review of a project's privacy effects and responses.\\
Prompt injection & Hostile instructions hidden in text that an AI reads, intended to make it misbehave or disclose information.\\
RAG (retrieval-augmented generation) & A design in which the AI fetches documents at question time and uses them to answer.\\
Subprocessor & Another company a vendor passes data to in order to deliver its service.\\
Tenant & An organisation's own account and settings within a shared cloud service.\\
Training & Changing an AI model using data, so that patterns from the data become part of the model.\\
\end{longtable}
\endgroup

\section{AI-use disclosure}
\label{app:ai}
ChatGPT using the GPT-6 Astra model (OpenAI), used on 20~September~2026, assisted with structuring the literature review, drafting prose and \LaTeX{} formatting. Earlier sessions also used Cursor Agent (Composer) and OpenAI Codex for the same kinds of drafting support. Claude (Anthropic), used between 1 and 10~October~2026, assisted with reviewing the manuscript, reducing repetition, rewriting prose for clarity, adding definitions and a glossary, checking bibliographic details of added sources, drafting the comparative-test dossiers, forms and expected routes, and acting as the simulated assessors (Claude Haiku) described in Section~\ref{sec:panel}. The authors selected the topic, research questions, university examples and decision rules; checked the public Lyrebird and ChatGPT Edu pages used in Tables~\ref{tab:lyrebird} and~\ref{tab:sydneyedu} on 21~September~2026 and rechecked both pages on 10~October~2026; and derived those routes from the gates and Equation~\eqref{eq:route} in Section~\ref{sec:framework}. The authors remain responsible for every claim in the paper. University of Sydney guidance requires this identification of the tool and of the assistance used \cite{sydneyuse}. This paper has not undergone journal peer review.

\end{document}